GENERALIZING FUZZY LOGIC PROBABILISTIC INFERENCES

SILVIO URSIC

URSIC COMPUTING
810 Ziegler Rd.
Madison, Wisconsin 53714

## 1. INTRODUCTION

The deduction problem we will consider can be loosely described as follows:
given the probabilities of some events, we wish to compute the probabilities of
some other events. The terms "event" and "probability" are to be understood as
having the meaning assigned to them in statistics. Probabilistic inference
rules have been shown to constitute a key ingredient of expert systems. This
report addresses the problem of generalizing the two fuzzy logic rules

$P(A \text{ AND } B) = \min(P(A), P(B))$ and
$P(A \text{ OR } B) = \max(P(A), P(B))$,                    (1)

for statistical events defined with boolean formulas more complex than a single
logical "AND" and a single logical "OR". It is well known, and readily shown
with small examples, that these two probabilistic inference rules, together
with $P(\text{not } A) = 1 - P(A)$, are not sufficient to accurately deduce probabilities
for collections of events defined in some arbitrary manner. The loss of
accuracy we incurr when we use only these two min-max rules depends on the
number of interacting events we are simultaneously considering. For two
events, the inference rules in (1) are exact. For small collections of events,
say four or five, they are in most occasions sufficient. When the number of
interacting events grows, their performance degrades. Sometimes they become
practically useless as the errors between the true probabilities and the
approximations they produce are too large to be acceptable.

Let A and B be two events from a universe U, and let $P(x)$ indicate a
probability measure on U. We then have:

$\max(0, P(A) + P(B) - 1) \leq P(A \& B) \leq \min(P(A), P(B))$,      (2)

where "&" is event intersection (a logical "AND"). For example, for
$P(A) = 0.4$ and $P(B) = 0.8$, from (1) we conclude that $0.2 \leq P(A \& B) \leq 0.4$.
The two inequalities in (2) for their simplicity, usefulness and beauty should
be considered a "classic". To this respect, see the problem as posed by
[Boole 1854] on page 298, and solved on page 299. The purpose of what follows
is to show how to obtain inequalities (2) and their generalizations for
arbitrarily defined events, as solutions to linear programs, while avoiding
the inherent exponential size in the number of necessary variables.
Conceptually, we follow Boole's plan for the problem, carried out with tools
and techniques totally unknown to him. One must realize that he did not even
have at his disposal duality theory for linear inequalities.

## 2. THE PRODUCT PARTITION

We will introduce with an example the techniques utilized to generalize the
fuzzy logic min-max rules for any collection of arbitrarily defined events.
These upper and lower bounds in (2) can be obtained with the following steps.

303

We start with the product partition probability simplex. It describes the fact that each event in the product partition generated by the collection of events under consideration is nonnegative. Also, as the events in the product partition are mutually disjoint, and their union is the whole universe, they must add up to one. We have (we indicate event complementation with "-"):

$$
\begin{aligned}
0 &\le P(A\&B) \\
0 &\le P(-A\&B) \\
0 &\le P(A\&-B) \\
0 &\le P(-A\&-B) \\
1 &= P(A\&B) + P(-A\&B) + P(A\&-B) + P(-A\&-B).
\end{aligned} \tag{3}
$$

Define the events of interest, in this case events A and B, as a union of events from the product partition. We have:

$$
\begin{aligned}
P(A) &= P(A\&B) \qquad\qquad\quad + P(A\&-B) \\
P(B) &= P(A\&B) + P(-A\&B).
\end{aligned} \tag{4}
$$

The system of linear equations and inequalities in (3) and (4) has six variables, namely $P(A)$, $P(B)$, $P(A\&B)$, $P(-A\&B)$, $P(A\&-B)$ and $P(-A\&-B)$. We eliminate the variables that do not interest us with a projection of the probability simplex on the linear subspace defined by $P(A)$, $P(B)$ and $P(A\&B)$. We obtain:

$$
\begin{aligned}
0 &\le P(A\&B) \\
0 &\le P(A) \qquad\quad - P(A\&B) \\
0 &\le P(B) - P(A\&B) \\
0 &\le 1 - P(A) - P(B) + P(A\&B)
\end{aligned} \tag{5}
$$

All the probabilities for events that do not interest us, namely $P(-A\&B)$, $P(A\&-B)$ and $P(-A\&-B)$, are thus eliminated. Each inequality in the projection (5) provides an upper or a lower bound for $P(A\&B)$, as a function of $P(A)$ and $P(B)$. Rewriting (5) to emphasize this fact we have:

$$
\begin{aligned}
0 &\le P(A\&B) \\
P(A\&B) &\le P(A) \\
P(A\&B) &\le P(B) \\
P(A) + P(B) - 1 &\le P(A\&B).
\end{aligned} \tag{6}
$$

These four inequalities can be written in a more compact form as upper and lower bounds to $P(A\&B)$, and we have formula (2).

The difficult part in the preceding steps is the computation of a description of the projection of the partition simplex on the subspace defined by the events of interest. The probability simplex is in a space whose dimension is an exponential in the number of events under consideration. To be able to effectively handle it we must first project it into some subspace of much smaller dimension. Each face of the projection will then provide one of the inequalities we are seeking. For example, to obtain (2) we need all four faces of the projection (5). In this small example the reduction in the number of variables is not significant. We reduced the number of events from six to three. With a larger number of events, the exponential growth of the product partition must be dealt with in an effective way, otherwise the entire procedure outlined in the example can be classified as a "thought experiment". We can apply it to ten events, but not one hundred.

### 3.  CHOICE OF A PROGRAMMING LANGUAGE FOR EVENT SPECIFICATION

The need for a programming language to specify events is essential.  The choice made here is to express events with boolean formulas in conjunctive normal form.  This choice has many beneficial properties.  It is, for example, the choice made with PROLOG. In our context, this choice limits the kind of linear subspaces in which we must project the probability simplex.  The limitation is drastic and simplifies matters consideraby.  The computation of the projection on the partition simplex is much easier as only subspaces of a specific form have to be manipulated.  This choice of boolean formulas in conjunctive normal form can be made totally transparent to a user input language.  Algorithms to convert a general description of an event in any suitable language to conjunctive normal form are well understood.  Hundreds of transformations to it have been catalogued in connection with the study of NP-complete problems.

The choice of boolean formulas in conjunctive normal form limits the subspaces on which to perform the projection of the partition simplex.  As a consequence it is possible to choose very convenient vector bases for them.  It can be shown that the collection of probabilities associated with clauses of a boolean formula in conjunctive normal form with no negations,  the monotone clauses,  form such a basis.  Consult [Ursic 84] for algebraic details and proofs.

### 4.  A SECOND EXAMPLE

The following example illustrates that other forms of  inequalities  arise besides  the ones in our first example and hence the min-max relations in (2) are not sufficient for the task at hand.  We essentially repeat our first example, now with three events.  We project the partition simplex on the three events A,  B,  and C onto the subspace defined by P(A), P(B), P(C),  P(A&B), P(A&C)  and P(B&C).  These six events form a basis for the linear  subspaces defined by all the boolean formulas in three variables and at most two literals per clause.  We obtain the inequalities (7), (8), (9) and (10):

$$
\begin{aligned}
1 - P(A) - P(B) + P(A\&B) &\geq 0 \\
P(A) \qquad\quad - P(A\&B) &\geq 0 \\
P(B) - P(A\&B) &\geq 0 \\
P(A\&B) &\geq 0
\end{aligned} \tag{7}
$$

$$
\begin{aligned}
1 - P(A) - P(C) + P(A\&C) &\geq 0 \\
P(A) \qquad\quad - P(A\&C) &\geq 0 \\
P(C) - P(A\&C) &\geq 0 \\
P(A\&C) &\geq 0
\end{aligned} \tag{8}
$$

$$
\begin{aligned}
1 - P(B) - P(C) + P(B\&C) &\geq 0 \\
P(B) \qquad\quad - P(B\&C) &\geq 0 \\
P(C) - P(B\&C) &\geq 0 \\
P(B\&C) &\geq 0
\end{aligned} \tag{9}
$$

$$
\begin{aligned}
1 - P(A) - P(B) - P(C) + P(A\&B) + P(A\&C) + P(B\&C) &\geq 0 \\
P(A) \qquad\qquad\qquad - P(A\&B) - P(A\&C) + P(B\&C) &\geq 0 \\
P(B) \qquad\qquad - P(A\&B) + P(A\&C) - P(B\&C) &\geq 0 \\
P(C) + P(A\&B) - P(A\&C) - P(B\&C) &\geq 0.
\end{aligned} \tag{10}
$$

The  first three groups of inequalities (7),  (8) and (9),  correspond  to inequalities that lead to the fuzzy logic rule (2). The last group (10) is of a different nature. This work was originally started with the aim of generalizing these inequalities for any N, not just N = 2 or N = 3. The next section is a sampler of the techniques used in this "hunt for inequalities". Additional information about the topic can be found in [Ursic 84].

## 5. PROJECTING THE PARTITION PROBABILITY SIMPLEX

The goal of this section is to show the type of techniques employed to produce inequalities, similar to the max and min relations for a single "AND", for any event arbitrarily defined with a boolean formula. The main tool to be used for this purpose are generating functions which list the characteristic functions defining the sets of truth assignments associated with clauses of a boolean formula in conjunctive normal form. The starting point is the following two by three matrix:

$$
A_x = \begin{array}{c c} & \begin{array}{c c c} \emptyset & x & \bar{x} \end{array} \\ & \left| \begin{array}{c c c} 1 & 1 & 0 \\ 1 & 0 & 1 \end{array} \right| & \begin{array}{c} F_x \\ T_x \end{array} \end{array} .
\tag{11}
$$

The labels x and $\bar{x}$ represent a boolean variable and its negation. The row labels $T_x$ and $F_x$ represent the two truth assignments to the boolean variable x. matrix $A_x$ can be interpreted as defining three characteristic functions, its three columns defining three sets on the two elements $T_x$ (True for x) and $F_x$ (False for x). The set labeled $\emptyset$ has as elements both $T_x$ and $F_x$. The three sets hence are the universal set $\emptyset$, and the two sets labeled x and $\bar{x}$. Next, consider the following tensor product:

$$
A_x \langle * \rangle A_y = \begin{array}{c} \begin{array}{c c c} \emptyset & x & \bar{x} \end{array} \\ \left| \begin{array}{c c c} 1 & 1 & 0 \\ 1 & 0 & 1 \end{array} \right| \begin{array}{c} F_x \\ T_x \end{array} \end{array} \langle * \rangle \begin{array}{c} \begin{array}{c c c} \emptyset & y & \bar{y} \end{array} \\ \left| \begin{array}{c c c} 1 & 1 & 0 \\ 1 & 0 & 1 \end{array} \right| \begin{array}{c} F_y \\ T_y \end{array} \end{array} = \begin{array}{c} \begin{array}{c c c | c c c | c c c} \emptyset & x & \bar{x} & y & xy & \overline{x}y & \bar{y} & x\bar{y} & \overline{xy} \end{array} \\ \left| \begin{array}{c c c | c c c | c c c} 1 & 1 & 0 & 1 & 1 & 0 & 0 & 0 & 0 \\ 1 & 0 & 1 & 1 & 0 & 1 & 0 & 0 & 0 \\ \hline 1 & 1 & 0 & 0 & 0 & 0 & 1 & 1 & 0 \\ 1 & 0 & 1 & 0 & 0 & 0 & 1 & 0 & 1 \end{array} \right| \begin{array}{c} F_x F_y \\ T_x F_y \\ F_x T_y \\ T_x T_y \end{array} \end{array}
\tag{12}
$$

Row and column labels for matrix (12) are sets whose elements are the row and column labels of $A_x$ and $A_y$. With our entries for matrix $A_x$, a set union is computed with a pairwise product of the entries in the corresponding characteristic functions. Hence the column labeled xy defines the characteristic function for x or y. Let us now apply to matrices (11) and (12) the techniques normally used in obtaining generating functions. We add a parameter t whose degree is used to sort the columns of matrix (12) by the size of their labels. First define two additional matrices $E_x$ (Empty label) and $L_x$ (Literal x).

$$
E_x = \begin{array}{c} \begin{array}{c} \emptyset \end{array} \\ \left| \begin{array}{c} 1 \\ 1 \end{array} \right| \begin{array}{c} F_x \\ T_x \end{array} \end{array} ; \quad L_x = \begin{array}{c} \begin{array}{c c} x & \bar{x} \end{array} \\ \left| \begin{array}{c c} 1 & 0 \\ 0 & 1 \end{array} \right| \begin{array}{c} F_x \\ T_x \end{array} \end{array} .
\tag{13}
$$

We now have $A_x = E_x \langle + \rangle L_x$, and the product in (12) can be rewritten with the t added. Separating terms we obtain:

$$(E_x \langle+\rangle L_x \langle*\rangle t) \langle*\rangle (E_y \langle+\rangle L_y \langle*\rangle t) =$$

$$
\begin{array}{c|c}
\emptyset \\
1 \\ 1 \\ 1 \\ 1
\end{array}
\begin{array}{c}
F_x F_x \\ T_x F_y \\ F_x T_y \\ T_x T_y
\end{array}
\langle+\rangle
\begin{array}{cccc}
x & \bar{x} & y & \bar{y} \\
1 & 0 & 1 & 0 \\
0 & 1 & 1 & 0 \\
1 & 0 & 0 & 1 \\
0 & 1 & 0 & 1
\end{array}
\begin{array}{c}
F_x F_y \\ T_x F_y \\ F_x T_y \\ T_x T_y
\end{array}
\langle*\rangle t \langle+\rangle
\begin{array}{cccc}
xy & \bar{x}y & x\bar{y} & \bar{x}\bar{y} \\
1 & 0 & 0 & 0 \\
0 & 1 & 0 & 0 \\
0 & 0 & 1 & 0 \\
0 & 0 & 0 & 1
\end{array}
\begin{array}{c}
F_x F_y \\ T_x F_y \\ F_x T_y \\ T_x T_y
\end{array}
\langle*\rangle t^2 .
$$

To obtain the generating function for the clauses of a boolean formula in conjunctive normal form in N variables and exactly i literals per clause, indicated with $C_{N,i}$, we write:

$$\langle*\rangle \ \ (E_i \langle+\rangle L_i \langle*\rangle t) = \langle+\rangle \ \ C_{N,i} \langle*\rangle t^i, \qquad (14)$$
$$1 \le i \le N \qquad\qquad 0 \le i \le N$$

where $L_i$ and $E_i$ are the matrices L and E indexed with $x_i$. Formula (14) defines a matrix with $3^N$ columns and $2^N$ rows. The terms with $t^i$, $0 \le i \le N$, select the matrices $C_{N,i}$, whose columns correspond to clauses with exactly $\bar{i}$ literals. As a consequence it is possible to use the binomial recursion [Ursic 84] to obtain an alternate definition of the matrices $C_{N,i}$. We have:

$$
C_{0,0} =
\begin{array}{c|c}
& \emptyset \\ \hline
1 & \emptyset;
\end{array}
\qquad \text{for } i \neq 0, \ C_{0,i} = \emptyset;
$$

$$
C_{N,i} = C_{N-1,i} \langle*\rangle
\begin{array}{c|c}
& \emptyset \\ \hline
1 & F_{x_i} \\
1 & T_{x_i}
\end{array}
\langle+\rangle \ C_{N-1,i-1} \langle*\rangle
\begin{array}{c|cc|c}
& x_i & \bar{x}_i & \\ \hline
& 1 & 0 & F_{x_i} \\
& 0 & 1 & T_{x_i}
\end{array}
\qquad (15)
$$

The recursive definition (15) is the wanted recursion. It generates matrices which define the meaning of the clauses of a boolean formula in conjunctive normal form in N variables and with exactly i literals per clause. With this algebraic machinery the pursuit of probability inequalities is much simplified. In fact, it becomes possible. An outline of the main results follows.

<u>Inequalities for a general collection of sets are defined by boolean symmetric functions with the uniform parity condition.</u>

Probability inequalities can be interpreted as boolean functions because they touch some subset of the $2^N$ vertices of the partition simplex. A boolean symmetric function [Seshu & Hohn 59], [Cunkle 63], [Arnold & Harrison 63]

$$s_{N,(a_0, \ a_1, \ \cdots, a_p)}$$

is completely defined by N, the number of boolean variables, and by a collection of at most N+1 integers $a_i$, with $0 \le a_i \le N$ and $0 \le i \le p \le N$. The integers $a_i$ are known as its defining constants. The uniform parity condition is a condition that becomes necessary for very natural reasons. It states that for $0 \le k \le N$ the quantity

$$(a_0 - k) * (a_1 - k) * (a_2 - k) * \ldots * (a_p - k),$$

for an inequality must have the same sign, either always positive or always negative. With this notation, inequalities in formulas (5) are given by the symmetric function $s_{2,(1,2)}$. The properties introduced by the fact that not all boolean symmetric functions give origin to valid inequalities explains many of the oddities present in the inequalities that arise in connection with combinatorial problems.

### In addition to the interchange and negation of boolean variables, a third symmetry is found to be present (the flip symmetry).

Having an inequality, we obtain additional ones with symmetries of the underlying polytopes. For example, the four inequalities in (5) are obtained from the symmetric function $s_{2,(1,2)}$ by negating boolean variables four times. We can also interchange, or permute, boolean variables. A third symmetry has also been found to be present. It fills a gap left by the permutation and negation of variables. The reader may verify that these two standard symmetries cover only sixteen out of the $4! = 24$ possible permutations of the truth assignments on two boolean variables. The other missing eight permutations give also origin to valid inequalities and are taken care by the flip symmetry. In essence, in this setting there is only one symmetry that permutes the truth table for two boolean variables in all the $4!$ ways. The negation of variables, the interchange of variables and the flip symmetry are all manifestations of this single symmmetry.

### Inequalities combine using Vandermonde convolution.

Partially symmetric functions, constructed by compounding symmetric functions with tensor products, give origin to valid inequalities as well. Much follows from this. The simplest consequence of these compound inequalities is that all the inequalities for a problem in N events are also inequalites for a problem in $N + i$, $i > 0$, events. Although one might think of this as an intuitively obvious statement, no formal proof of this escalator property was previously known.

### 6. SOME FINAL COMMENTS

An outline of how to use the techniques presented here for the problem of computing event probabilities is as follows.

Define all the events of interest with boolean formulas in conjunctive normal form. This can always be done in an efficient way (i.e. in polynomial time). For example, the boolean formula (A & B) is already in this form. It is a single clause with two literals. Similarly, the formula (A OR B) is a boolean formula with two clauses in one literal each. All the transformations developed to show that some problem is NP-complete can be directly used for this purpose.

Obtain upper and lower bounds on the probabilities of the events of interest with small linear programs. The relation min( P(A), P(B) ) can be considered such a linear program. Typically, in order to handle five to ten events simultaneously one must be able to solve linear programs in a few dozen variables. The inequalities to be used will be of two types. The first type is problem dependent. They describe the events under consideration and are in correspondence with the clauses of the boolean formulas that define them. The second type of inequality is fixed. They are in correspondence with boolean symmetric functions with the uniform parity condition and reflect the projection of the partition simplex on our smaller subspace. This projection is performed algebraically and reduces the dimension of the subspace to be considered from exponential to polynomial. Consult [Ursic 84] for algebraic details. This second type of inequalities need not be permanently stored in working memory. They are generated as needed.

Many peculiarities of these inequalities can be used to simplify the task of using them during the computation of the probability of some event. Boolean symmetric functions have many exploitable properties. The final product resembles more a combinatorial search than a simplex method computation of minima and maxima. The precision with which we compute probabilities depends on how many of these linear inequalities associated with symmetric functions we are willing to consider in the optimization phase. The ultimate precision obtainable depends also on how many events we wish to simultaneously consider at each step in the computations. The rules in (2) consider two events at a time. Experience with available code indicates that, after an initial stage in which we obtain our probabilities with some precision relatively quickly, the available algorithms settle to a stabler asymptotic behavior. In this asymptotic stage, a doubling of the precision seems to require at least a doubling of the computing time, all the way to exponential time for absolute precision. There is considerable heuristic evidence that this exponential behavior cannot be bested.

The philosophical meaning of this is that the problem of computing probabilities (considering $P \neq NP$) is a computationally open ended one. Finer and finer discriminations between very similar statistical events will require larger and larger amounts of computing time (or more powerful computers). It gives an additional meaning to the phrase "Let me think a little longer about it".

The practical meaning of this state of affairs is that we can trade precision for computing time in a remarkably controlled way. An effort is currently under way to quantify more precisely this trade-off between precision and computing time. Many additional implementation details about the techniques outlined here can be found in [Ursic 86].